\documentclass{article}

\usepackage[preprint]{neurips_2025}

\usepackage{adjustbox}
\usepackage{makecell}
\usepackage{multicol}
\usepackage{multirow}
\usepackage{hhline}
\usepackage{booktabs}
\usepackage{color}
\usepackage{xcolor}
\usepackage{amssymb}
\usepackage{subfigure} 
\usepackage{amsmath}
\usepackage{CJK}
\usepackage{enumitem}
\usepackage{natbib}
\usepackage{tabularx}

\newcolumntype{L}[1]{>{\raggedright\arraybackslash}p{#1}}
\newcolumntype{C}[1]{>{\centering\arraybackslash}p{#1}}
\newcolumntype{R}[1]{>{\raggedleft\arraybackslash}p{#1}}

\definecolor{purple}{RGB}{128, 0, 128}

\usepackage[utf8]{inputenc} 
\usepackage[T1]{fontenc}    
\usepackage{hyperref}       
\usepackage{url}            
\usepackage{booktabs}       
\usepackage{amsfonts}       
\usepackage{nicefrac}       
\usepackage{microtype}      
\usepackage{xcolor}         

\usepackage{longtable} 
\DeclareSymbolFont{extraup}{U}{zavm}{m}{n}
\DeclareMathSymbol{\varheart}{\mathalpha}{extraup}{86}
\DeclareMathSymbol{\vardiamond}{\mathalpha}{extraup}{87}
\DeclareMathSymbol{\varclubsuit}{\mathalpha}{extraup}{88}

\title{DNAZEN: Enhanced Gene Sequence Representations via Mixed Granularities of Coding Units}

\author{
    Lei Mao$^{\clubsuit}$, \hspace{0.1cm}
    Yuanhe Tian$^{\varheart}$, \hspace{0.1cm}
    Yan Song$^{\spadesuit*}$
     \\
    $^{\clubsuit}$Origin Omics \hspace{0.1cm}
    $^{\varheart}$University of Washington
    \\
    $^{\spadesuit}$University of Science and Technology of China \\
    $^{\clubsuit}$\texttt{maolei@originomics-ai.com} 
     \vspace{0.1cm}
    $^{\varheart}$\texttt{yhtian@uw.edu} \hspace{0.1cm}
    $^{\spadesuit}$\texttt{clksong@gmail.com}  \\
}

\begin{document}

\renewcommand{\thefootnote}{\fnsymbol{footnote}}
\footnotetext[1]{Corresponding author.}

\renewcommand{\thefootnote}{\arabic{footnote}}

\maketitle

\begin{abstract}

Genome modeling conventionally treats gene sequence as a language, reflecting its \textcolor{black}{structured motifs} and long-range dependencies analogous to linguistic units and organization principles
such as words and syntax.
Recent studies utilize advanced neural networks, ranging from convolutional and recurrent models to Transformer-based models, to capture contextual information of gene sequence, 
with the primary goal of obtaining effective gene sequence representations and thus enhance the models' understanding of various running gene samples.
However, these approaches often directly apply language modeling techniques to gene sequences and do not fully consider the intrinsic information organization in them, where they do not consider how units at different granularities contribute to representation.
In this paper, we propose DNAZEN, an enhanced genomic representation framework designed to learn from 
various granularities in gene sequences, including small polymers and G-grams that are combinations of several contiguous polymers.
Specifically, we extract the G-grams from large-scale genomic corpora through an unsupervised approach to construct the G-gram vocabulary, which is used to provide G-grams in the learning process of DNA sequences through dynamically matching from running gene samples.
A Transformer-based G-gram encoder is also proposed and 
the matched G-grams are fed into it to compute their representations and integrated into \textcolor{black}{the encoder for basic unit (E4BU)},
which is responsible for encoding small units and maintaining the learning and inference process.
To further enhance the learning process, we propose whole G-gram masking to train DNAZEN, where the model largely favors the selection of each entire G-gram to mask rather than an ordinary masking mechanism performed on basic units.
Experiments on the genome understanding evaluation (GUE) benchmark show that DNAZEN outperforms representative models on various downstream tasks such as promoter detection, transcription factor prediction, splice site prediction, etc.
The results confirm that G-grams with an appropriate learning process improve genomic representation through generalization ability to capture broader bioinformatic signals while maintaining high efficiency in genome modeling.\footnote{Code and models are available at \url{https://github.com/oomics/dnazen}.}

\end{abstract}

\section{Introduction}

Gene sequences offer rich information underpinning biological functions, and the analysis upon them supports breakthroughs in biology and medicine over years \citep{gill1994identification,he2010graphene,corcoran2018application}.
Similar to linear chain signals such as natural language,
gene sequences share common characteristics to them so that genome modeling is able to perform such studies as if they are processing a special type of language \citep{ji_linguistics_1999,yoon_gene_2002,searls_language_2002}.
To have an accurate understanding of the information carried by gene sequences, similar to text, it requires an appropriate processing method, where precise encoding of units at various granularities (e.g., nucleotide bases, polymers, oligonucleotides, etc) is critically important.

Existing studies 
\citep{ji2021dnabert,nguyen2023hyenadna,liu2024exploring,schiff2024caduceus,nguyen2024sequence,dalla2024nucleotide} utilize various types of model structures, such as convolutional neural networks (CNNs) \citep{lecun1998gradient}, recurrent neural networks (RNNs) \citep{rumelhart1986learning}, and Transformer \citep{vaswani2017attention}, etc.
Among different models, most existing studies adopt Transformer \citep{ji2021dnabert,zhang2023dnagpt,zhou2023dnabert,li_applications_2023,rozowsky_en-tex_2023,choi_transformer_2023} as their foundation models because it demonstrates its effectiveness in language modeling for various scenarios \citep{devlin2019bert,brown2020language}.
In language modeling, the basic units of language (tokens) are crucial, since variants such as characters, subwords, and words carry different semantic information and thus produce distinct learning outcomes.
For example, in English, characters (letters) have poor semantic expressiveness and yield unsatisfactory learning results, while words struggle to cover novel terms and weaken generalization, so subwords, which both convey meaning independently and support generalization, become the dominant approach \cite{devlin2019bert,brown2020language}.
Similarly, for gene sequences, choosing the appropriate unit of representation for learning is equally important.
Although it is straightforward to directly use single nucleotides (i.e., \texttt{A}, \texttt{C}, \texttt{G}, \texttt{T}) as the molecular units, they lead to an issue of small alphabet, resulting in the problem of lacking structural modeling ability. 
Therefore, genome modeling approaches with different architectures often utilize a tokenizer to segment the sequence into small tokens, which in most cases are larger than single nucleotides and serve as the basic building blocks for the learning process of gene sequence representations \citep{choi_transformer_2023,Dahiya2023From,Estabrook2023Predicting,Ma2023Disease-gene,Morimoto2023Regional}.
For example, ``\textit{k-mer}'' is a widely used tokenization mechanism that uses a fixed number of nucleotide bases for each unit \citep{wen2014k};
some studies \citep{zhou2023dnabert} employ byte pair encoding (BPE) \citep{sennrich2015neural} to segment DNA sequences into dynamically grouped nucleotide bases,
and shows its superiority in learning gene sequence representations in many cases \citep{zhou2023dnabert}.
There are also studies that combine different tokenization strategies \citep{wu2025generator}, leading to better performance on some downstream tasks than the models that rely on a single tokenization strategy.
Although these studies achieve a promising effect in bioinformation tasks, they pay limited attention to larger coding units,
which might play essential roles in understanding important information carried by gene sequences \citep{peng1992long}, because a sequence must be sufficiently long to encapsulate an independent and distinct biological function.
Still, one may argue that genome evolution is driven by single-nucleotide mutations \citep{lamason2005slc24a5,kamberov2013modeling}, each of which effectively creates a new gene sequence with possible different functions,
so that a particular coding scheme may be greatly affected by such mutations, thus leading to the inefficient encoding problem.
Moreover, understanding biomolecular interactions potentially requires ultimate fine-grained sequence resolution, since critical functional sites may consist of only one or a few specific nucleotide bases \citep{stamatoyannopoulos1972molecular,enattah2002identification}.
Therefore, the encoding of gene sequences actually faces a dilemma, as it must simultaneously accommodate the processing of both large and small coding units while in view of the necessity of them.
To this end, it is essential to develop an improved learning framework for gene sequence representation that effectively handles such coding scheme at multiple granularities, particularly by enhancing the processing of large units,
so as to address the limitations of previous approaches that primarily focused on polymers or groups of a few nucleotide bases.

\begin{figure}[t]
    \centering
    \includegraphics[width=1.0\linewidth, trim=0 20 0 0]{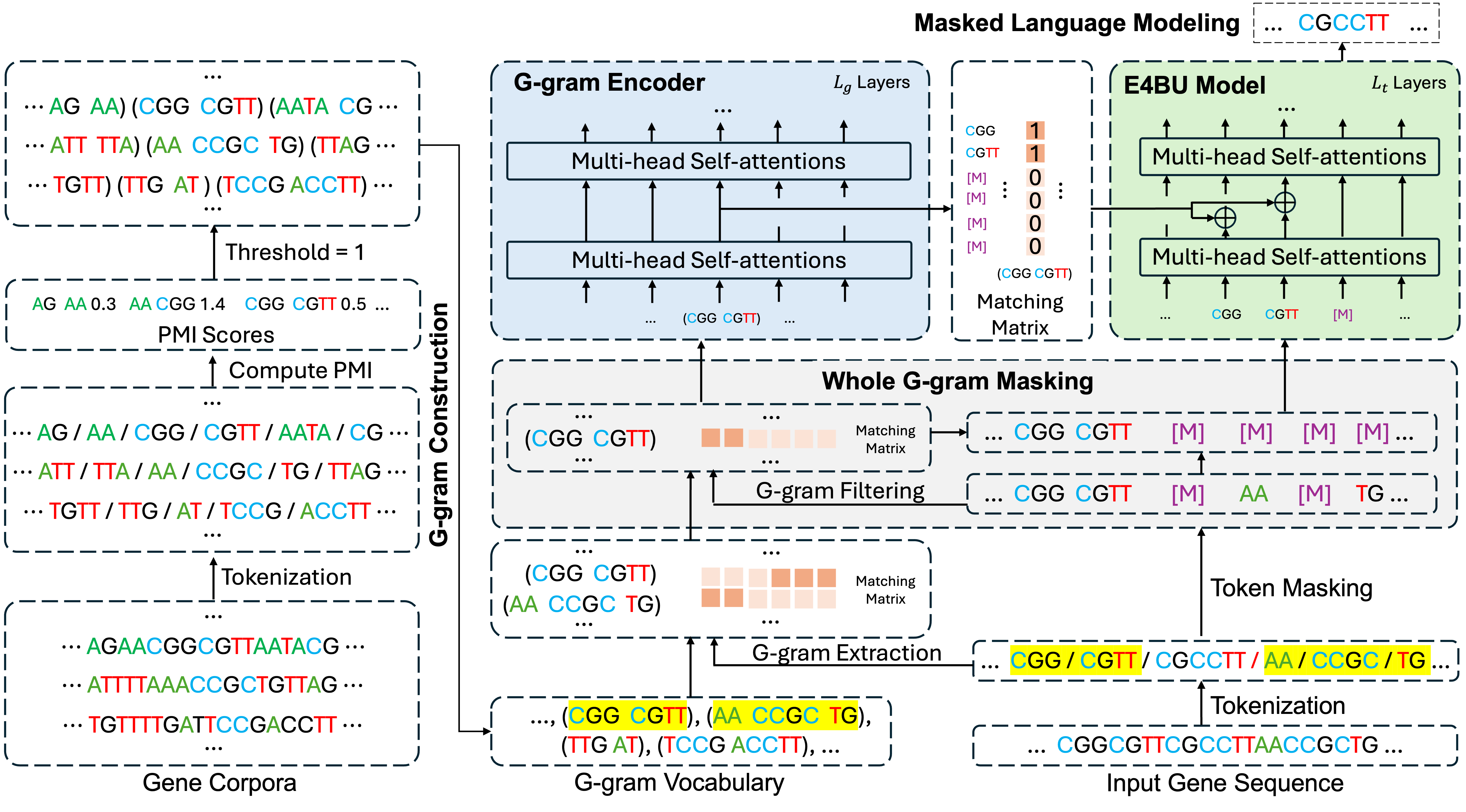}
    \caption{\textcolor{black}{
    The overview of DNAZEN.
    The left part presents the G-gram construction process, where the G-grams are obtained from large-scale gene corpora through an unsupervised approach named pointwise mutual information (PMI).
    The constructed G-gram vocabulary, the extracted G-grams, the matching matrix between the input tokens and the G-grams, as well as the G-gram encoder are shown in the middle of the figure.
    The original and tokenized input gene sequence, as well as the \textcolor{black}{E4BU} to enhance the gene representation with G-grams and to perform masked language modeling, are presented at the right part of the figure.
    The middle-right part is the whole G-gram masking used in pre-training to learn information of different granularities.
    }}
    \label{fig:model}
    \vspace{-0.2cm}
\end{figure}

In this paper, we propose DNAZEN, a Transformer-based foundation framework for genome modeling with the enhancement of G-grams\footnote{\textcolor{black}{We use the term ``G-gram'' to show our respect to the term ``n-gram'' in natural language processing that conveys a similar idea.}}, which are dynamically extracted large coding units for gene sequences
by combining smaller continuous segments, such as BPE of nucleotide bases.
Specifically,
we extract G-gram from large-scale genomic corpora by computing pointwise mutual information (PMI) between adjacent base units and merging those with PMI above a threshold into large-grained segments, which are collected to construct a G-gram vocabulary.
For each input gene sequence, we tokenize it and extract the G-grams appearing in both the input and the vocabulary from the tokenized sequence.
We utilize the Transformer-based G-gram encoder and \textcolor{black}{an encoder for basic unit (E4BU)} to encode the extracted G-grams and input tokens, respectively, where the G-gram and token representations are computed in each Transformer layer.
\textcolor{black}{In each layer,} the G-gram representations are integrated with the representations of \textcolor{black}{the associated tokens} in the \textcolor{black}{E4BU} to form an enriched representation that contains multi-granular information.
Moreover, we also propose whole G-gram masking (WGM) to further enhance the learning process of G-grams, where the masking process is performed on both \textcolor{black}{token} and G-gram levels, 
where DNAZEN prefers to mask each entire G-gram rather than the individual tokens in the standard marking mechanism,
which allows DNAZEN to learn genomic information at multiple granularities.
Experiments on multi-species genome benchmarks reveal that DNAZEN improves the performance of various downstream tasks, demonstrating its effectiveness via the superior performance over strong baselines.

\section{The DNAZEN}

To better capture broader biological information within larger coding units,
DNAZEN enhances the representation learning capabilities of conventional Transformer encoders by explicitly incorporating G-gram into the standard masked language modeling process jointly with the learning of smaller coding units.
The overall architecture of DNAZEN is presented in Figure \ref{fig:model}.
Specifically, DNAZEN integrates a G-gram extraction procedure that constructs a G-gram vocabulary in an unsupervised manner, where the resulting G-grams are used to match those large units in any input gene sequences.
DNAZEN encodes both the small and large units in the input and integrates the representations of the G-grams into those of the small units to compute the enhanced representations.
We propose a whole G-gram masking (WGM) strategy to train DNAZEN to learn biomedical information in both small and large granularities.
Finally, we perform fine-tuning on downstream tasks with labeled data following the standard pre-training and fine-tuning paradigm. 
In the following text, we present the G-gram construction, G-gram encoding, representation enhancement with Genseqs, and training and inference with G-grams in detail.

\subsection{G-gram Construction}
\label{sec:G-gram extraction}

DNA sequences contain biological signals that span multiple nucleotides.
Although existing studies \citep{zhou2023dnabert} use BPE to segment input sequences into tokens and achieve promising results, these tokens often break contiguous biological patterns into shorter coding units, \textcolor{black}{or split gene sequences with independent biomedical functions into several pieces,} 
so that hurting models’ ability to properly learning from broader informative contexts.
By enlarging such context and enriching input features,
DNAZEN considers more basic coding units, such as BPE, at a time through G-grams, which
are constructed beforehand according to those small units.
\textcolor{black}{
In doing so, G-grams are conservative sequences extracted from diverse genomic regions from a biological perspective, which makes them similar to motifs that also represent conserved gene structures.
Meanwhile, G-grams are also used to gauge the granularity of sequence elements that serve as building blocks of biological function.
Consider that one may argue that motifs have different lengths compared with G-grams, we utilize a statistic-based algorithm to construct G-grams and determine their length so that they are statistically meaningful units that contribute to the understanding of gene sequence.
}

In constructing the G-gram vocabulary $\mathcal{V}$, we firstly \textcolor{black}{utilize a gene sequence tokenizer to segment the sequences in large genomic corpora into small units, i.e., tokens.
We employ a statistic-based approach to measure the informational tightness between continuous small units; if their connection is strong, we combine those neighboring units into larger ones.
}
\textcolor{black}{Specifically}, we adopt pointwise mutual information (PMI) to perform G-gram extraction under the following principle.
Given two adjacent units $x'$ and $x''$, their PMI score is computed by
\begin{equation}
\setlength\abovedisplayskip{8pt}
\setlength\belowdisplayskip{8pt}
\text{PMI}(x', x'') = \log \frac{p(x'x'')}{p(x')p(x'')},
\end{equation}
where $p$ represents the probability of the occurrence $x'x''$ and individual tokens $x'$ and $x''$ within the corpus, respectively.
A higher PMI score indicates a significant co-occurrence frequency, suggesting the two adjacent units should have a strong binding and potentially be considered as a part of G-gram.
Therefore,
we compute through all gene sequences for the PMI scores between any adjacent units and use a threshold $\theta$ to guide weather a delimiter should be inserted at certain positions.
For instance, if an input sequence is segmented as ``\texttt{ATA} / \texttt{CGGT} / \texttt{TGTA} / \texttt{GGTT} / \texttt{AGGA}'',
\textcolor{black}{
and the PMI scores between ``\texttt{ATA} / \texttt{CGGT}'', ``\texttt{CGGT} / \texttt{TGTA}'', ``\texttt{TGTA} / \texttt{GGTT}'', and ``\texttt{GGTT} / \texttt{AGGA}'' are 0.7, 1.3, 1.1, and 0.1, respectively.
Then, given a threshold of $\theta=1$, we find the pairs whose PMI score is lower than the threshold and insert a delimiter between the two tokens of the pair.
In this example, it leads to a delimiter between the tokens in the first and last pairs, and thus results in a G-gram of ``\texttt{CGGT} / \texttt{TGTA} / \texttt{GGTT}''.
}
Accordingly, the aforementioned procedure is performed over
the entire gene sequence dataset and all G-grams are collected and cleaned to form the vocabulary $\mathcal{V}$.

\subsection{G-gram Encoding}
\label{sec:genseq encoding}

The encoding of G-grams is not individually done but
collectively performed in each running gene sequence, i.e., we use each gene sequence as the object of encoding and all G-grams in it are encoded at the same time.
In doing so, we firstly use $\mathcal{V}$ to extract all G-grams recorded in $\mathcal{V}$ from each gene sequence.
That is,
given an input sequence $\mathcal{X}=x_1x_2\dots x_n \dots x_N$, we searches for G-grams recorded in $\mathcal{V}$ and collects them to get a G-grams list $\mathcal{C} = [c_1 \cdots c_t \cdots c_T]$ for $\mathcal{X}$, where G-grams with overlapping are allowed.\footnote{
Notably, such G-gram searching and collection process is performed dynamically on the running input gene sequence at both training and inference stages.
}

Then,
DNAZEN utilizes a multi-layer Transformer \textcolor{black}{(the number of layers is denoted as $L_{g}$)} to encode G-grams, as shown on the \textcolor{black}{middle part} of Figure \ref{fig:model},
which utilizes standard multi-head self-attention mechanisms to model interactions among G-grams,
so that the model is able to dynamically emphasize biologically salient G-grams based on their contextual relevance within all G-grams in a collection.
Note that, standard Transformer assume that all inputs should present in a sequential order, while such G-grams do not have this characteristic when they are treated as normal input to Transformer, positional embeddings are then omitted in the process of encoding them, allowing the model to treat all G-grams equally and focus purely on their biological significance regardless of their positions.
Each extracted G-gram from the sequence is represented by an embedding vector initialized from a G-gram embedding matrix and fed into this encoder, \textcolor{black}{where the embedding of the $t$-th G-gram is denoted as $\mathbf{e}^g_t$.}
Formally, we use \textcolor{black}{$\mathbf{\mu}_t^{(l-1)}$ to denote the input representation of the $t$-th G-gram at layer $l$ (the input representation of the first layer is the G-gram embedding $\mathbf{e}^g_t$).
Each $\mathbf{\mu}_t^{(l-1)}$ is then recomputed via multi-head self‐attention over the set $\{\mathbf{\mu}_1^{(l-1)},\dots,\mathbf{\mu}_T^{(l-1)}\}$, producing enriched representation $\{\mathbf{\mu}_1^{(l)},\dots,\mathbf{\mu}_T^{(l)}\}$ that capture contextual information by
\begin{equation}
    \mathbf{\mu}_1^{(l)},\dots,\mathbf{\mu}_T^{(l)} = f^{(l)}_g(\mathbf{\mu}_1^{(l-1)} \cdots \mathbf{\mu}_T^{(l-1)})
\end{equation}
where $f^{(l)}_g$ denotes the multi-head attentions at the $l$-th layer of the G-gram encoder. 
}

\subsection{Representation Enhancement with G-grams}

Although G-grams carry large-grained contextual information of DNA sequence, it is difficult to directly learn their representations with only using the G-gram encoder owning the the following reasons.
First, large-grained units produce sparse representations, where most G-grams may appear rarely in the training set to support stable embedding learning. 
Second, large-grained units lead to overfitting, since the model tends to memorize these rare G-grams rather than capture general sequence patterns. 
Third, large-grained units introduce noise by incorporating irrelevant variations from sequencing errors, which obscure genuine biological signals.
Therefore, we employ a \textcolor{black}{E4BU} to encode the input tokens and conduct specific learning tasks as standard language models do.
Specifically, as shown in Figure \ref{fig:model}, this \textcolor{black}{E4BU} is a Transformer-based architecture that firstly maps each token to an embedding and then encodes them layer by layer using a Transformer with $L_t$ layers ($L_t>L_g$).
We denote the representation of the $n$-th token $x_n$ at layer $l$ as $\mathbf{\nu}_n^{*(l)}$, and the final layer outputs the representation $\mathbf{\nu}_n^{*(L_t)}$.
To facilitate learning both small- and large-grained information, we feed tokens and G-grams into DNAZEN and encode each with its own encoder; at the same time, we use a standard language modeling paradigm to mask and recover the input, thereby capturing both levels of information.
Therefore, we fuse G-gram representations from the G-gram encoder into the \textcolor{black}{E4BU model}, enabling the benefits from its conventionally used learning objectives and the ability to leverage existing pre-trained parameters,
so as to borrow its effectiveness in applying to training and downstream fine-tuning.

In detail, we construct a matching matrix $\mathbf{M} \in \{0,1\}^{N \times T}$ after the G-gram extraction process, where $N$ is the sequence length measured by the basic units used for G-gram construction, and $T$ is the number of matched G-grams.
Each entry $m_{n,t}$ indicates whether the $n$-th unit belongs to the $t$-th G-gram by
\begin{equation}
m_{n,t} =
\begin{cases}
1, & \text{if token } x_n \text{ belongs to G-gram } c_t \\
0, & \text{otherwise}
\end{cases}
\end{equation}
which connects the information from basic units to broader contextual evidence provided by the G-grams.
Afterwards, using the matching matrix, DNAZEN integrates the G-gram representations obtained from the G-gram encoder with representations of basic units generated by the \textcolor{black}{E4BU} model at its each layer from bottom to top
to progressively enhance the overall representation of the entire gene sequence with all its G-grams' information.
Specifically, the enhanced representation $\mathbf{\nu}_n^{*(l)}$ for each unit $x_n$ at layer $l$ is computed by aggregating the embeddings of all associated G-grams by
\begin{equation} \label{eq:combine}
\mathbf{\nu}_n^{*(l)} = \mathbf{\nu}_n^{(l)} + \sum_{t} m_{n,t} \mathbf{\mu}_t^{(l)}
\end{equation}
where 
$\mathbf{\nu}_n^{(l)}$ is the original representation for $x_n$ from the \textcolor{black}{E4BU} model
\textcolor{black}{and the integration process is performed for all layers with $l\in[1, L_g]$}.
Through this integration, DNAZEN effectively integrates longer contiguous genomic patterns into the representation learning process, enriching its capability in contextual modeling beyond the scope of polymers or nucleotide bases.

\subsection{Training and Inference with G-grams}

DNAZEN follows the standard pre-training procedure from existing studies \cite{zhou2023dnabert,zhang2023dnagpt}.
\textcolor{black}{
During each training step, a subset of the input tokens are randomly selected to be ``masked'' (e.g., be replaced by a special token named ``\texttt{[MASK]}'') so that the encoder has no direct access to masked token. 
The model is then optimized to recover the original tokens from the surrounding unmasked context, thereby learning meaningful contextual representations.
}
Inspired by the success of whole-word masking (WWM) in langauage modeling \cite{sun2019ernie,cui2021pre}, DNAZEN proposes a similar strategy ``whole G-gram masking'' to enhance the learning objective, masked language modeling, during pre-training.
The WGM
consists of three stages.
First, we perform the standard unit masking where we randomly select some of them to be masked with a certain probability.
Second, if a unit is found to appear in one or more particular G-grams, all units belonging to these G-grams are simultaneously masked.\footnote{
To prevent data leakage affecting the learning objective, these G-grams in the G-gram list $\mathcal{C}$ are also temporally masked and are not modeled by the G-gram encoder for the current input gene sequence.}
Third, following the conventional masking procedure, DNAZEN replaces 80\% of the masked units with a special ``[MASK]'' symbol, substitutes 10\% of them with randomly chosen units, and leaves the remaining 10\% unchanged. 
In doing so,
this masking approach allows the model to effectively leverage information from both large and small units according to their contextual environments.
\textcolor{black}{For inference, we follow the convention in language modeling, where the original input without any masking is fed into the model to process all tokens and G-grams to obtain the input representations.}

\begin{table}[t]
\centering
\caption{Summary of tasks and statistics in the GUE benchmarks.}
\scalebox{1.0}{
\begin{tabular}{lcccc}
\toprule
Species & Task & Num. Datasets & Num. Classes & Sequence Length \\
\midrule
Human & Core Promoter Detection & 3 & 2 & 70 \\
Human & Transcription Factor Prediction & 5 & 2 & 100 \\
Human & Promoter Detection & 3 & 2 & 300 \\
Human & Splice Site Prediction & 1 & 3 & 400 \\
Mouse & Transcription Factor Prediction & 5 & 2 & 100 \\
Yeast & Epigenetic Marks Prediction & 10 & 2 & 500 \\
Virus & Covid Variant Classification & 1 & 9 & 1000 \\
\bottomrule
\end{tabular}
}
\label{tab:tasks}
\end{table}

\section{Experiments}

\subsection{Datasets and Tasks}

We use the same pre-training and fine-tuning datasets as that used in DNABERT-2 \citep{ji2021dnabert}.
The pre-training data include the human genome and a multi-species genome collection, which contains 35 billion nucleotide bases in total. 
For evaluation, we adopt the Genome Understanding Evaluation (GUE) benchmark, which covers various genomic analysis tasks, including core promoter detection (CPD), transcription factor prediction (TFP), promoter detection (PD), splice site prediction (SSP), epigenetic marks prediction (EMP), and covid variant classification (CVC), respectively.
Table \ref{tab:tasks} summarizes the main tasks along with the number of correspondent datasets, the number of classes, and the sequence length\footnote{\textcolor{black}{In GUE, all sequences in a dataset have the same length.}} (in terms of the number of nucleotide bases in each instance) for each task.
We follow \cite{zhou2023dnabert} to process the data and split the GUE benchmark into train/dev/test sets.

\begin{table}[tp]
    \centering
    \caption{Performance comparison across species and tasks. 
The best results are highlighted in \textbf{boldface}; the DNAZEN results that are higher than DNABERT-2 under the same settings are marked by \underline{underlines}.
The baselines include DNABERT \citep{ji2021dnabert}, DNABERT-2 \citep{zhou2023dnabert}, and Nucleotide Transformer \citep{dalla2024nucleotide} with the results obtained from \cite{zhou2023dnabert}.
\textcolor{black}{
``Avg. G-gram per Case'' denotes the average number of G-gram in each test instance.
}
``$^\dag$'' marks the models that are further pre-trained on the GUE datasets.
}
    \label{tab:all results}
    \setlength{\tabcolsep}{4pt}
    \begin{tabularx}{\textwidth}{@{\extracolsep{\fill}}l|cccccc}
        \toprule
        Task & \multicolumn{6}{c}{Epigenetic Marks Prediction} \\
        \midrule
        Dataset & H3    & H3K14ac & H3K36me3 & H3K4me1 & H3K4me2 & H3K4me3 \\
        \midrule
        Avg. G-gram per Case & 11.5 & 12.7 & 12.6 & 12.2 & 12.4 & 12.7 \\
        \midrule
        DNABERT    & 74.15 & 42.07 & 48.49 & 42.95 & 31.34 & 28.92 \\
        DNABERT-2 & 78.27 & 52.57 & 56.88 & 50.52 & 31.13 & 36.27 \\
        DNABERT-2$^\dag$ & 80.17 & 57.42 & 61.90 & 53.00 & 39.89 & 41.20 \\
        Nucleotide Transformer    & 78.77 & 56.20 & 61.99 & 55.30 & 36.49 & 40.34 \\
        \midrule
        DNAZEN & \underline{\textbf{81.72}} & \underline{55.59} & \underline{\textbf{64.33}} & \underline{57.51} & 27.76 & \underline{40.22} \\
        DNAZEN$^\dag$ & \underline{81.09} & \underline{\textbf{59.28}} & \underline{{63.19}} & \underline{\textbf{58.14}} & \underline{\textbf{39.73}} & \underline{\textbf{42.84}} \\
        \bottomrule
    \end{tabularx}
    \begin{tabularx}{\textwidth}{@{\extracolsep{\fill}}l|cccc|ccc}
        \toprule
         Task & \multicolumn{4}{c|}{Epigenetic Marks Prediction} & \multicolumn{3}{c}{Promoter Detection} \\
        \midrule
        Dataset & H3K79me3 & H3K9ac & H4 & H4ac & all & notata & tata \\
        \midrule
        Avg. G-gram per Case & 12.2 & 12.5 & 11.0 & 12.7 & 7.7 & 7.9 & 5.8 \\
        \midrule
        DNABERT   & 60.12 & 50.48 & 78.27 & 38.60 & 90.44 & 93.61 & 69.83 \\
        DNABERT-2 & 67.39 & 55.63 & 80.71 & 50.43 & 86.77 & 94.27 & 71.59 \\
        DNABERT-2$^\dag$ & 65.46 & 57.07 & 81.86 & 50.35 & 88.31 & \textbf{94.34} & 68.79 \\
        Nucleotide Transformer   & 64.70 & 56.01 & 81.67 & 49.13 & \textbf{91.01} & 94.00 & \textbf{79.43} \\
        \midrule
        DNAZEN & \underline{67.60} & \underline{\textbf{60.12}} & \underline{80.71} & \underline{51.32} & 85.38 & 93.52 & 68.67\\
        DNAZEN$^\dag$ & \underline{\textbf{67.78}} & \underline{{59.27}} & \underline{\textbf{82.24}} & \underline{\textbf{53.15}} & 87.03 & 93.80 & \underline{69.84} \\
        \bottomrule
    \end{tabularx}
    \begin{tabularx}{\textwidth}{@{\extracolsep{\fill}}l | ccccc | ccc}
        \toprule
         Tasks & \multicolumn{5}{c|}{Transcription Factor Prediction (Human)} & \multicolumn{3}{c}{Core Promoter Detection} \\
        \midrule
        Dataset & 0 & 1 & 2 & 3 & 4 & all & notata & tata \\
        \midrule
        Avg. G-gram per Case & 4.0 & 3.4 & 3.4 & 1.7 & 2.8 & 1.8 & 1.9 & 0.8 \\
        \midrule
        DNABERT   & 67.95 & 70.90 & 60.51 & 53.03 & 69.76 & \textbf{70.92} & 69.82 & 78.15 \\
        DNABERT-2 & \textbf{71.99} & \textbf{76.06} & 66.52 & 58.54 & 77.43 & 69.37 & 68.04 & 74.17 \\
        DNABERT-2$^\dag$ & 69.12 & 71.87 & 62.96 & 55.35 & 74.94 & 67.50 & 69.53 & 76.18 \\
        Nucleotide Transformer   & 66.64 & 70.28 & 58.72 & 51.65 & 69.34 & 70.33 & \textbf{71.58} & 72.97 \\
        \midrule
        DNAZEN & 66.82 & {73.89} & 65.51 & \underline{{59.13}} &  \underline{\textbf{80.83}} & 64.93 & 67.44 & \underline{77.96} \\
        DNAZEN$^\dag$ & 68.35 & \underline{\textbf{75.07}} & \underline{\textbf{66.66}} & \underline{\textbf{60.68}} & \underline{\textbf{78.81}} & \underline{67.62} & 68.21 & \underline{\textbf{79.15}} \\
        \bottomrule
    \end{tabularx}
    \begin{tabularx}{\textwidth}{@{\extracolsep{\fill}}l | ccccc | c | c}
        \toprule
        Tasks & \multicolumn{5}{c|}{Transcription Factor Prediction (Mouse)} & \multicolumn{1}{c|}{Virus} & \multicolumn{1}{c}{Splice} \\
        \midrule
        Dataset & 0 & 1 & 2 & 3 & 4 & Covid & Reconstruct \\
        \midrule
        Avg. G-gram per Case & 1.8 & 1.7 & 1.4 & 0.6 & 2.2 & 126.8 & 8.8 \\
        \midrule
        DNABERT   & 42.31 & 79.10 & 69.90 & 55.40 & 41.97 & 62.23 & 84.14 \\
        DNABERT-2 & 56.76 & 84.77 & 79.32 & 66.47 & \textbf{52.66} & 71.02 & 84.99 \\
        DNABERT-2$^\dag$ & \textbf{64.23} & \textbf{86.28} & 81.28 & 73.49 & 50.80 & 68.49 & 85.93 \\
        Nucleotide Transformer   & 63.31 & 83.76 & 71.52 & 69.44 & 47.07 & \textbf{73.04} & 89.35 \\
        \midrule
        DNAZEN & \underline{62.23} & 84.21 & \underline{{82.16}} & \underline{\textbf{76.37}} &  49.66 & 60.10 & \underline{\textbf{89.60}} \\
        DNAZEN$^\dag$ & 62.25 & 85.17 & \underline{\textbf{83.02}} & \underline{75.12} & \underline{52.45} & 66.54 & \underline{89.35} \\
        \bottomrule
    \end{tabularx}
\end{table}

\subsection{Implementation Details}

\textcolor{black}{To construct the G-gram vocabulary for DNAZEN modeling, we utilize the BPE tokenizer} used for DNABERT-2 \citep{zhou2023dnabert} with 4,096 vocab size.
The max sequence length of the \textcolor{black}{E4BU} is set to 512 tokens.
For G-gram extraction, we use the tokenizer to segment all pre-training data and set the PMI threshold to 2.
We keep the G-grams whose token-based length is between 2 and 5 (including 2 and 5) and obtain a G-gram vocabulary with 163K distinct G-grams.
In DNAZEN, the G-gram encoder utilizes 6 layers of Transformer and 
\textcolor{black}{with 768 hidden size, the 
\textcolor{black}{E4BU} follows the same architecture as DNABERT-2 \citep{zhou2023dnabert} with 12-layer Transformer with 768 hidden size.
We train DNAZEN on eight NVIDIA H800 Tensor Core GPUs.
}

For pre-training and fine-tuning, we \textcolor{black}{follow the standard paradigm with the details illustrated as follows.}
During pre-training, DNAZEN employs a masked language modeling objective with a mask ratio of 15\%.
The model is trained on the combined \textcolor{black}{human and multi-species genome dataset}
using the same splits as DNABERT-2.
\textcolor{black}{
The \textcolor{black}{E4BU} is initialized by the pre-trained parameters of DNABERT-2, and the G-gram encoder is randomly initialized, where the parameters in both encoders are updated in training.
The learning rate is set to $5\times10^{-5}$.
We train the model for three epochs with a batch size of 1,024. 
In addition, we follow DNABERT-2 to also have a setting that further trains DNAZEN on the GUE data for another three epochs.
In the fine-tuning stage, we tune the hyperparameters based on the model's performance on the development set and evaluate it on the test set.
We follow existing studies \citep{ji2021dnabert,zhou2023dnabert} in using the Matthews correlation coefficient (MCC) as the evaluation metric, where higher values indicate better performance.
Each experiment runs with three different random seeds, and the average test set performance is reported. 
}

\begin{table}[t]
    \centering
    \caption{Performance of DNAZEN with and without whole G-gram masking (WGM).
}
    \label{tab:wgm}
    \setlength{\tabcolsep}{4pt}
    \begin{tabularx}{\textwidth}{@{\extracolsep{\fill}}l|cccccc}
        \toprule
        Task & \multicolumn{6}{c}{Epigenetic Marks Prediction} \\
        \midrule
        Dataset & H3    & H3K14ac & H3K36me3 & H3K4me1 & H3K4me2 & H3K4me3 \\
        \midrule
        DNAZEN w/o WGM & \textbf{81.86} & \textbf{60.26} & 61.29 & 57.81 & 38.27 & \textbf{44.16} \\
        DNAZEN w/ WGM & 81.09 & 59.28 & \textbf{63.19} & \textbf{58.14} & \textbf{39.73} & {42.48} \\
        \bottomrule
    \end{tabularx}
    \begin{tabularx}{\textwidth}{@{\extracolsep{\fill}}l|cccc|ccc}
        \toprule
         Task & \multicolumn{4}{c|}{Epigenetic Marks Prediction} & \multicolumn{3}{c}{Promoter Detection} \\
        \midrule
        Dataset & H3K79me3 & H3K9ac & H4 & H4ac & all & notata & tata \\
        \midrule
        DNAZEN w/o WGM & \textbf{62.36} & \textbf{60.39} & \textbf{82.37} & 51.73 & 85.73 & 93.32 & 67.06 \\
        DNAZEN w/ WGM & {{67.78}} & {{59.27}} & {{82.24}} & \textbf{53.15} & \textbf{87.03} & \textbf{93.80} & {\textbf{69.84}} \\
        \bottomrule
    \end{tabularx}
    \begin{tabularx}{\textwidth}{@{\extracolsep{\fill}}l | ccccc | ccc}
        \toprule
         Tasks & \multicolumn{5}{c|}{Transcription Factor Prediction (Human)} & \multicolumn{3}{c}{Core Promoter Detection} \\
        \midrule
        Dataset & 0 & 1 & 2 & 3 & 4 & all & notata & tata \\
        \midrule
        DNAZEN w/o WGM & 66.78 & 72.23 & 64.48 & 59.11 & 77.00 & 65.90 & 67.38 & 77.48 \\
        DNAZEN w/ WGM & \textbf{68.35} & {\textbf{75.07}} & {\textbf{66.66}} & {\textbf{60.68}} & {\textbf{78.81}} & \textbf{67.62} & \textbf{68.21} & {\textbf{79.15}} \\
        \bottomrule
    \end{tabularx}
    \begin{tabularx}{\textwidth}{@{\extracolsep{\fill}}l | ccccc | c | c}
        \toprule
        Tasks & \multicolumn{5}{c|}{Transcription Factor Prediction (Mouse)} & \multicolumn{1}{c|}{Virus} & \multicolumn{1}{c}{Splice} \\
        \midrule
        Dataset & 0 & 1 & 2 & 3 & 4 & Covid & Reconstruct \\
        \midrule
        DNAZEN w/o WGM & 59.86 & 83.97 & \textbf{85.22} & 74.72 & 50.29 & 62.38 & 87.20 \\
        DNAZEN w/ WGM & \textbf{62.25} & \textbf{85.17} & {{83.02}} & {\textbf{75.12}} & {\textbf{52.45}} & \textbf{66.54} & {\textbf{89.35}} \\
        \bottomrule
    \end{tabularx}
\end{table}

\section{Results and Analysis}

\subsection{Overall Results}

\textcolor{black}{
We evaluate the performance of DNAZEN by comparing it with several strong baseline models, including DNABERT \citep{ji2021dnabert}, DNABERT-2 \citep{zhou2023dnabert}, and Nucleotide Transformer \citep{dalla2024nucleotide}. 
DNABERT is designed explicitly for genomic sequences, utilizing overlapping k-mer tokenization methods. 
Nucleotide Transformer further enhanced genomic modeling by scaling up the size of both model parameters and training data, demonstrating promising performance across various DNA analysis tasks.
Table \ref{tab:all results} summarizes the performance of all models across multiple genomic prediction tasks and species, including Epigenetic Marks Prediction (EMP), Transcription Factor Prediction (TFP), Covid Variant Classification (CVC), Promoter Detection (PD), Core Promoter Detection (CPD), and Splice Site Prediction (SSP).
The average number of G-gram for each case is also reported for reference.
The models marked by ``\dag'' are further pre-trained on the GUE dataset.
There are several observations.
}

\textcolor{black}{
First, compared to DNAZEN that is not pre-trained on GUE, DNAZEN$\dag$ achieves better performance on 22 out of 28 datasets, demonstrating the benefit of pre-training on downstream tasks.
Second, DNAZEN$\dag$ consistently outperforms DNABERT-2$\dag$ (both are further pre-trained on GUE datasets) across most datasets (i.e., 21 out of 28 datasets), where a similar trend is observed when comparing DNAZEN and DNABERT-2.
DNAZEN also outperforms many state-of-the-art performance set by Nucleotide Transformer. 
These performance gains indicate the effectiveness of incorporating G-gram representations, suggesting that DNAZEN better captures long-range biological patterns and contextual dependencies essential for genomic sequence understanding.
Third, we note that in general, DNAZEN obtains higher improvements over existing studies with more numbers of G-gram in the test instance. 
This observation demonstrates the effectiveness of modeling large-grained information for improving the understanding of DNA sequences.
}

\textcolor{black}{
To have a deeper understanding of the performance, we discuss the performance of different models on various tasks.
Overall, DNAZEN performs well in straightforward protein–DNA interaction tasks by modeling conserved sequence patterns effectively.
For epigenetic mark prediction (EMP), which links DNA sequences to histone modifications, DNAZEN achieves state-of-the-art performance by leveraging G-grams.
In transcription factor binding site identification (TFP), DNAZEN robustly captures sequence–protein interactions across human and mouse genomes.
Splice site prediction (SSP) benefits from DNAZEN’s ability to detect conserved splicing signals, yielding improved accuracy over baseline models.
In promoter detection (PD), where complex multi-protein transcription initiation complexes act on 300-nucleotide-unit inputs, NT achieves overall better performance, which is attributed to its larger size.
In core promoter detection (CPD) with 70-nucleotide-unit inputs, DNAZEN achieves comparable performance to DNABERT and NT by learning from tokens and G-grams in different granularities.
For Covid variant classification (CVC) tasks relying on subtle mutational patterns beyond conserved regions, DNAZEN’s emphasis on common sequences offers fewer benefits.
}

\begin{table}[t]
    \centering
    \caption{\textcolor{black}{Performance of DNAZEN pre-trained from scratch or from pre-trained DNABERT-2, which are marked by ``(R)'' and ``(P)'', respectively.}
}
    \label{tab:init}
    \setlength{\tabcolsep}{4pt}
    \begin{tabularx}{\textwidth}{@{\extracolsep{\fill}}l|cccccc}
        \toprule
        Task & \multicolumn{6}{c}{Epigenetic Marks Prediction} \\
        \midrule
        Dataset & H3    & H3K14ac & H3K36me3 & H3K4me1 & H3K4me2 & H3K4me3 \\
        \midrule
        DNAZEN (R) & 80.91 & \textbf{59.94} & 62.18 & \textbf{59.33} & \textbf{40.83} & \textbf{43.61} \\
        DNAZEN (P) & {\textbf{81.09}} & {{59.28}} & {\textbf{63.19}} & {{58.14}} & {{39.73}} & {{42.48}} \\
        \bottomrule
    \end{tabularx}
    \begin{tabularx}{\textwidth}{@{\extracolsep{\fill}}l|cccc|ccc}
        \toprule
         Task & \multicolumn{4}{c|}{Epigenetic Marks Prediction} & \multicolumn{3}{c}{Promoter Detection} \\
        \midrule
        Dataset & H3K79me3 & H3K9ac & H4 & H4ac & all & notata & tata \\
        \midrule
        DNAZEN (R) & \textbf{68.15} & \textbf{60.08} & 81.00 & \textbf{54.39} & 83.14 & 91.26 & 67.67 \\
        DNAZEN (P) & {{67.78}} & {{59.27}} & {\textbf{82.24}} & {{53.15}} & \textbf{87.03} & \textbf{93.80} & {\textbf{69.84}} \\
        \bottomrule
    \end{tabularx}
    \begin{tabularx}{\textwidth}{@{\extracolsep{\fill}}l | ccccc | ccc}
        \toprule
         Tasks & \multicolumn{5}{c|}{Transcription Factor Prediction (Human)} & \multicolumn{3}{c}{Core Promoter Detection} \\
        \midrule
        Dataset & 0 & 1 & 2 & 3 & 4 & all & notata & tata \\
        \midrule
        DNAZEN (R) & 67.20 & 68.86 & 58.26 & 52.80 & 74.84 & 59.47 & 64.04 & 75.34 \\
        DNAZEN (P) & \textbf{68.35} & {\textbf{75.07}} & {\textbf{66.66}} & {\textbf{60.68}} & {\textbf{78.81}} & \textbf{67.62} & \textbf{68.21} & {\textbf{79.15}} \\
        \bottomrule
    \end{tabularx}
    \begin{tabularx}{\textwidth}{@{\extracolsep{\fill}}l | ccccc | c | c}
        \toprule
        Tasks & \multicolumn{5}{c|}{Transcription Factor Prediction (Mouse)} & \multicolumn{1}{c|}{Virus} & \multicolumn{1}{c}{Splice} \\
        \midrule
        Dataset & 0 & 1 & 2 & 3 & 4 & Covid & Reconstruct \\
        \midrule
        DNAZEN (R) & 51.61 & 80.08 & 81.76 & 70.83 & 43.62 & 54.58 & 84.68 \\
        DNAZEN (P) & \textbf{62.25} & \textbf{85.17} & {\textbf{83.02}} & {\textbf{75.12}} & {\textbf{52.45}} & \textbf{66.54} & {\textbf{89.35}} \\
        \bottomrule
    \end{tabularx}
\end{table}

\subsection{Effect of Model Initialization Approach}

\textcolor{black}{
Table \ref{tab:init} presents performance comparisons of DNAZEN initialized from scratch (R) versus from the pre-trained DNABERT-2 model (P).
In most EMP tasks, random initialization yields slightly better or comparable performance.
In complex tasks such as transcription factor prediction and core promoter detection, pre-trained initialization further enhances model performance stability.
Notably, in mouse transcription factor prediction and splice site reconstruction tasks, pre-trained initialization significantly improves model generalizability.
These results demonstrate that DNAZEN is able to achieve strong performance even without a pre-trained model and further leverage existing pre-trained models to enhance its performance.
}

\subsection{Effect of Whole G-gram Masking}

\textcolor{black}{
To explore the effect of whole G-gram masking (WGM) on model performance, we pre-train DNAZEN without WGM following exactly the same setting as our full model and fine-tune it on downstream tasks.
We report the results (i.e., MCC) of DNAZEN without WGM on all datasets in Table \ref{tab:wgm}, where the results of DNAZEN with WGM are also reported for reference.
It is observed that, among 28 datasets, DNAZEN with WGM achieves better performance than DNAZEN without WGM on 21 datasets, which indicates the effectiveness of WGM for helping DNAZEN to learn G-gram information.
We attribute these gains to WGM’s ability to force the model to reconstruct entire G-gram, thereby enhancing the learning of co-occurrence patterns and long-range dependencies.  
}

\begin{figure}[t]
    \centering
    \includegraphics[width=0.9\linewidth, trim=0 15 0 0]{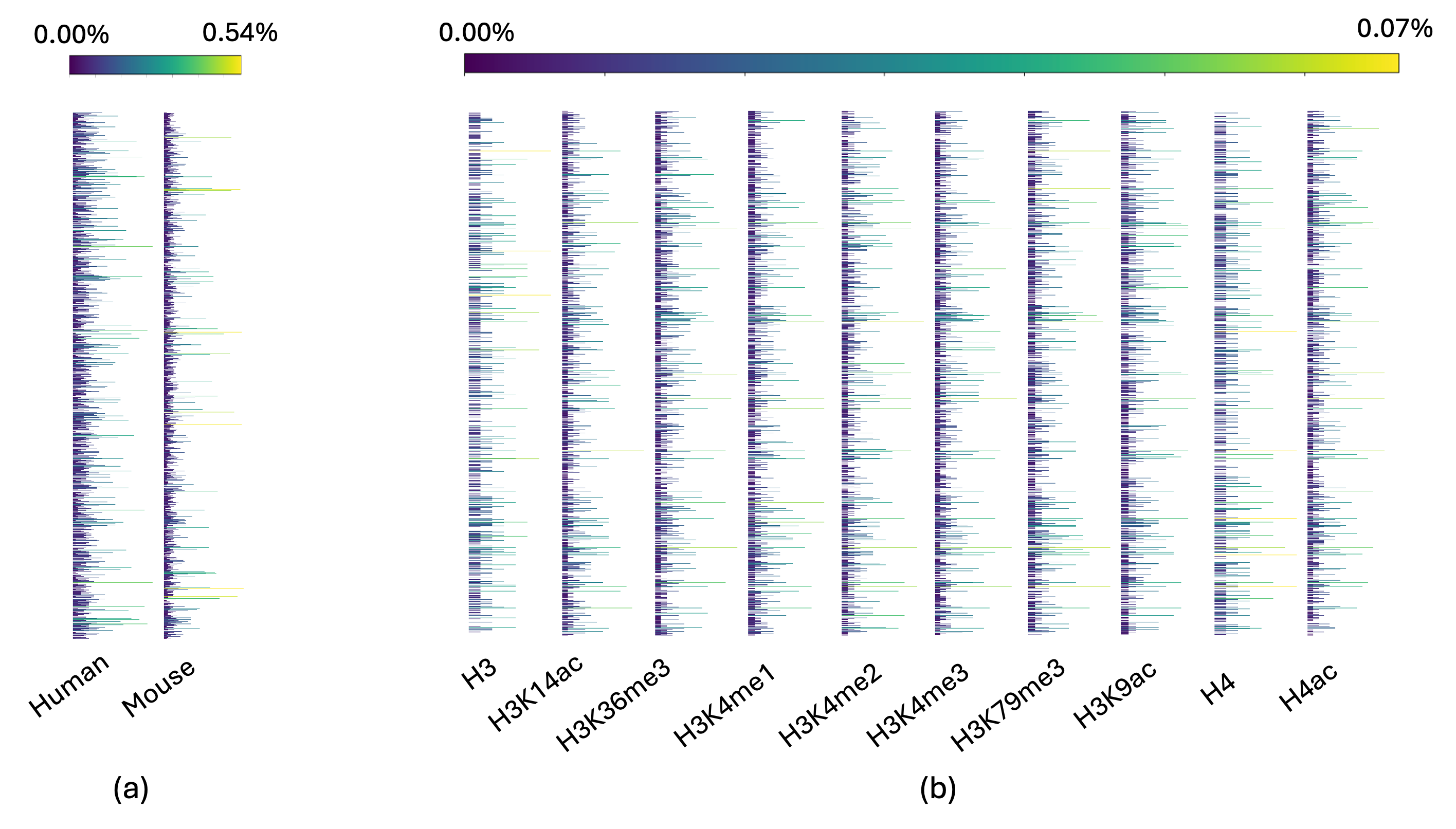}
    \caption{\textcolor{black}{
    The G-gram distribution in different datasets. Figure (a) presents the distribution on the transcription factor prediction tasks for human and mouse; Figure (b) shows the distribution on ten datasets of epigenetic marks prediction.
    For both figures, the y-axis denotes different G-grams (which are not presented for better visualization), and the length of the bar and the color intensity represent each G-gram’s frequency divided by the total number of G-grams in that dataset.
    }}
\label{fig:distribution}
\end{figure}

\subsection{Visualization of G-gram Distributions}

\textcolor{black}{
To further illustrate DNAZEN’s ability to capture both species- and dataset-specific sequence signatures, we visualize in Figure \ref{fig:distribution} the normalized frequency distributions of G-grams across multiple datasets. 
In Figure \ref{fig:distribution}(a), we aggregate the transcription factor prediction tasks for human and mouse and plot the relative frequency of the G-grams (normalized by the total number of G-grams in each dataset). 
Despite the identical prediction objective, the two species shows markedly different enrichment profiles: certain G-grams appear with high relative frequency in human but are rare in mouse, and vice versa. 
This clear inter-species divergence demonstrates that our dynamically extracted G-gram vocabulary effectively encodes species-specific regulatory motifs.
Meanwhile, in Figure \ref{fig:distribution}(b), we compare the distribution of G-grams across ten yeast epigenetic marks prediction datasets. 
Even within the same organism and task category, individual histone modification datasets present distinct hotspots in the heatmap: 
specific G-grams are highly enriched in one mark but depleted in others. 
Such intra-species, inter-dataset variability underscores DNAZEN’s capacity to learn fine-grained, dataset-specific sequence features under a consistent biological context.
}

\subsection{Case Study}

\begin{figure}[t]
    \centering
    \includegraphics[width=0.9\linewidth, trim=0 10 0 0]{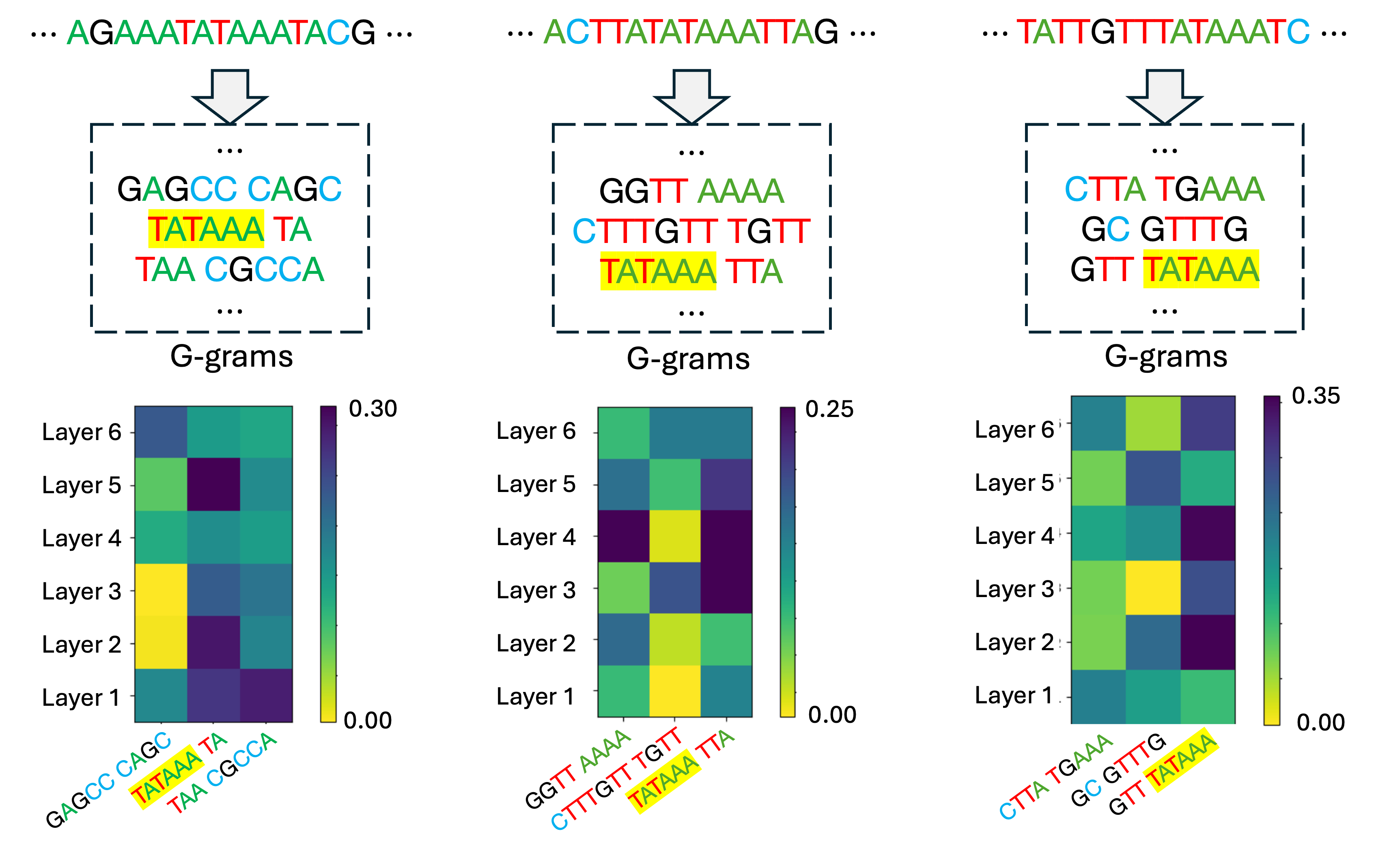}
    \caption{
    \textcolor{black}{
    Three cases to illustrate the effect of G-grams in helping DNAZEN to make correct predictions on the promoter detection task (the ones with \texttt{TATA} boxes).
    A part of the input DNA sequence is presented on the top, and the bottom shows some G-grams extracted from the input sequence.
    The G-grams with \texttt{TATA} box are highlighted by the dashed orange boxes. 
    }}
\label{fig:case}
\end{figure}

\textcolor{black}{
Figure \ref{fig:case} presents three promoter detection cases and the average attention weight for each G-gram in different layers of the G-gram encoder, where darker colors refer to higher weights. 
In the three instances, each gene sequence contains a canonical \texttt{TATA} box\footnote{\textcolor{black}{A typical \texttt{TATA} box is one from the sequences \texttt{TATA(A/T)A(A/T)}.}}, a fundamental cis-regulatory element that recruits the transcription pre-initiation complex in its genomic context.  
For all cases, our approach is able to extract G-grams covering the core ``\texttt{TATAAA}'' motif and its contextual tokens form a dedicated G-gram.  
Meanwhile, the G-gram encoder is able to assign higher weights to the G-grams containing the \texttt{TATA} box, which allows the G-gram contribute more to the gene understanding.
By encoding these G-grams, DNAZEN is able to accentuate the critical role of the \texttt{TATA} box, thereby making the model’s decision process more interpretable.  
Consequently, DNAZEN leverages these enriched G-gram representations to reveal how G-grams and their context contribute to predictions, demonstrating that dynamic segment-level modeling enhances interpretability.
}

\section{Related Work}

\textcolor{black}{
Genomic language modeling treats DNA sequences as a form of natural language, encoding critical biological signals \citep{ji_linguistics_1999,yoon_gene_2002,searls_language_2002,alipanahi_predicting_2015,zhou_predicting_2015,quang_danq_2016,kim_multi-omics_2016}. 
Early studies \citep{shen2018recurrent,liu2019detection,nguyen2023hyenadna} utilize CNN \citep{lecun1998gradient} and RNNs \citep{rumelhart1986learning} to analyze DNA sequences.
Most recent researchers apply Transformer-based architectures to generate numerical representations from DNA sequences, uncovering hidden genomic patterns \cite{ji2021dnabert,zhang2023dnagpt,zhou2023dnabert,li_applications_2023,rozowsky_en-tex_2023,choi_transformer_2023}.
%
%
Most existing approaches typically employ fixed-length segmentation methods, such as k-mer tokenization, dividing DNA sequences into overlapping, equal-length segments \citep{ji2021dnabert,zeng_4mcpred-mtl_2021,he_nucleic_2021,lopez_nucleotide_2023,zhang2023dnagpt}. 
Recent advancements incorporate subword tokenization strategies borrowed from natural language processing, notably byte-pair encoding (BPE), enabling dynamic merging of frequent nucleotide pairs into variable-length tokens \citep{zhou2023dnabert}. 
This dynamic, frequency-based tokenization enhances models' ability to represent complex genetic structures and reduces sequence length for more efficient processing \cite{zhou2023dnabert}.
There are some studies \citep{wu2025generator} that combine different types of tokenization approaches so as to take advantage of them.
}
\textcolor{black}{
Our approach differs from existing studies by integrating an extra G-gram extraction component that recovers longer contiguous segments from the genomic sequence. 
The G-gram encoder processes these segments and merges the resulting embeddings with the token representations in the \textcolor{black}{E4BU}, thereby improving the model's understanding of extended biological patterns.
}

\section{Conclusion}

\textcolor{black}{
In this paper, we propose DNAZEN, a genome foundation model that extends existing Transformer-based models by integrating extra G-gram representations. 
Our approach leverages two separate encoders to model both tokens and G-grams in the gene sequence, where the G-gram representations are incorporated into token representations to enhance the gene representations.
In addition, we propose whole G-gram masking that enables DNAZEN to learn from the G-gram in the input sequence in the pre-training process.
Experiments on multi-species benchmarks demonstrate the effectiveness of our approach, which outperforms strong baselines and existing studies on many tasks.
Further analyses further reveals the robustness of DNAZEN to train from scratch or leverage existing pre-trained models. 
We also demonstrate that DNAZEN is able to leverage important segments for DNA sequence representation, thereby enhancing model interpretability for downstream tasks.
}

\bibliographystyle{unsrtnat}
\bibliography{reference}

\newpage

\appendix

\section{The Statistics of the Extracted G-gram in GUE Benchmark}

\textcolor{black}{
In Table \ref{tab:detailed_statistics}, we present the total number of G-gram, the number of distinct G-grams, and the average number of G-grams for each instance in the training, development, and test sets of GUE.
}

\begin{longtable}[t]{lllr r r r}
  \caption{The statistics of G-gram in the training, development, and test sets of GUE, where the total and distinct number of G-grams, as well as the average number of G-gram per case, are reported.}
  \label{tab:detailed_statistics}\\
  \toprule
  Species & Task & Dataset & Split & Total G-gram & Distinct G-gram & Avg.\ G-gram \\
  \midrule
  \endfirsthead

  \multicolumn{7}{c}{\textit{Continued from previous page}}\\
  \toprule
  Species & Task & Dataset & Split & Total G-gram & Distinct G-gram & Avg.\ G-gram \\
  \midrule
  \endhead

  \midrule
  \multicolumn{7}{r}{\textit{Continued on next page}}\\
  \endfoot

  \bottomrule
  \endlastfoot

  \multirow[c]{36}{*}{Human} & \multirow[c]{9}{*}{CPD}
    & \multirow{3}{*}{all}    & train &  92,801   & 16,935  & 2.0 \\
  &                         &        & dev   &  11,290   &  5,234  & 1.9 \\
  &                         &        & test  &  10,622   &  4,636  & 1.8 \\
  \cmidrule(lr){3-7}
  &                         & \multirow{3}{*}{notata} & train &  85,350   & 15,762  & 2.0 \\
  &                         &        & dev   &  10,391   &  4,943  & 2.0 \\
  &                         &        & test  &  10,127   &  4,513  & 1.9 \\
  \cmidrule(lr){3-7}
  &                         & \multirow{3}{*}{tata}   & train &   7,442   &  4,477  & 1.5 \\
  &                         &        & dev   &     932   &    802  & 1.5 \\
  &                         &        & test  &     507   &    415  & 0.8 \\
  \cmidrule(lr){2-7}

  & \multirow[c]{15}{*}{TFP}
    & \multirow{3}{*}{0}      & train & 126,719   & 19,232  & 3.9 \\
  &                           &        & dev   &   3,870   &  2,507  & 3.9 \\
  &                           &        & test  &   3,963   &  2,396  & 4.0 \\
  \cmidrule(lr){3-7}
  &                           & \multirow{3}{*}{1}      & train & 112,350   & 19,114  & 3.7 \\
  &                           &        & dev   &   3,640   &  2,540  & 3.6 \\
  &                           &        & test  &   3,379   &  2,291  & 3.4 \\
  \cmidrule(lr){3-7}
  &                           & \multirow{3}{*}{2}      & train &  76,772   & 14,880  & 4.0 \\
  &                           &        & dev   &   3,983   &  2,486  & 4.0 \\
  &                           &        & test  &   3,435   &  2,132  & 3.4 \\
  \cmidrule(lr){3-7}
  &                           & \multirow{3}{*}{3}      & train &  59,129   & 20,017  & 2.2 \\
  &                           &        & dev   &   2,255   &  1,935  & 2.3 \\
  &                           &        & test  &   1,659   &  1,410  & 1.7 \\
  \cmidrule(lr){3-7}
  &                           & \multirow{3}{*}{4}      & train &  62,022   & 15,039  & 3.3 \\
  &                           &        & dev   &   3,057   &  2,250  & 3.1 \\
  &                           &        & test  &   2,830   &  1,947  & 2.8 \\
  \cmidrule(lr){2-7}

  & \multirow[c]{9}{*}{PD}
    & \multirow{3}{*}{all}    & train & 385,441   & 32,117  & 8.1 \\
  &                         &        & dev   &  47,102   & 12,811  & 8.0 \\
  &                         &        & test  &  45,323   & 11,253  & 7.7 \\
  \cmidrule(lr){3-7}
  &                         & \multirow{3}{*}{notata} & train & 351,331   & 30,155  & 8.3 \\
  &                         &        & dev   &  43,109   & 12,072  & 8.1 \\
  &                         &        & test  &  41,778   & 10,828  & 7.9 \\
  \cmidrule(lr){3-7}
  &                         & \multirow{3}{*}{tata}   & train &  33,957   & 12,497  & 6.9 \\
  &                         &        & dev   &   4,170   &  2,929  & 6.8 \\
  &                         &        & test  &   3,557   &  2,286  & 5.8 \\
  \cmidrule(lr){2-7}

  & \multirow[c]{3}{*}{SSD}
    & \multirow{3}{*}{ssd}    & train & 327,259   & 39,950  & 9.0 \\
  &                         &        & dev   &  40,219   & 16,122  & 8.8 \\
  &                         &        & test  &  40,097   & 15,596  & 8.8 \\
  \midrule

  \multirow[c]{15}{*}{Mouse} & \multirow[c]{15}{*}{TFP}
    & \multirow{3}{*}{0}      & train &  15,332   &  6,233  & 2.4 \\
  &                           &        & dev   &   1,917   &  1,436  & 2.4 \\
  &                           &        & test  &   1,466   &  1,004  & 1.8 \\
  \cmidrule(lr){3-7}
  &                           & \multirow{3}{*}{1}      & train &  99,746   & 17,337  & 1.9 \\
  &                           &        & dev   &  12,035   &  5,199  & 1.8 \\
  &                           &        & test  &  11,768   &  4,476  & 1.7 \\
  \cmidrule(lr){3-7}
  &                           & \multirow{3}{*}{2}      & train &   5,029   &  2,325  & 1.9 \\
  &                           &        & dev   &     639   &    444  & 1.9 \\
  &                           &        & test  &     468   &    232  & 1.4 \\
  \cmidrule(lr){3-7}
  &                           & \multirow{3}{*}{3}      & train &   2,536   &  1,956  & 1.3 \\
  &                           &        & dev   &     314   &    276  & 1.3 \\
  &                           &        & test  &     151   &    116  & 0.6 \\
  \cmidrule(lr){3-7}
  &                           & \multirow{3}{*}{4}      & train &  39,017   &  9,792  & 2.6 \\
  &                           &        & dev   &   4,749   &  2,741  & 2.5 \\
  &                           &        & test  &   4,128   &  2,211  & 2.2 \\
  \midrule

  \multirow[c]{30}{*}{Yeast} & \multirow[c]{30}{*}{EMP}
    & \multirow{3}{*}{H3}      & train & 180,697   & 74,080  & 15.1 \\
  &                           &        & dev   &  20,995   & 15,226  & 14.0 \\
  &                           &        & test  &  17,161   & 11,808  & 11.5 \\
  \cmidrule(lr){3-7}
  &                           & \multirow{3}{*}{H3K14ac} & train & 371,764   &110,712  & 14.1 \\
  &                           &        & dev   &  43,524   & 26,913  & 13.2 \\
  &                           &        & test  &  42,124   & 25,802  & 12.7 \\
  \cmidrule(lr){3-7}
  &                           & \multirow{3}{*}{H3K36me3} & train & 386,935   &112,645  & 13.9 \\
  &                           &        & dev   &  44,848   & 27,222  & 12.9 \\
  &                           &        & test  &  43,771   & 26,365  & 12.6 \\
  \cmidrule(lr){3-7}
  &                           & \multirow{3}{*}{H3K4me1}  & train & 352,291   &108,724  & 13.9 \\
  &                           &        & dev   &  40,244   & 24,698  & 12.7 \\
  &                           &        & test  &  38,567   & 23,771  & 12.2 \\
  \cmidrule(lr){3-7}
  &                           & \multirow{3}{*}{H3K4me2}  & train & 340,972   &107,546  & 13.9 \\
  &                           &        & dev   &  39,514   & 24,603  & 12.9 \\
  &                           &        & test  &  37,943   & 23,682  & 12.4 \\
  \cmidrule(lr){3-7}
  &                           & \multirow{3}{*}{H3K4me3}  & train & 407,861   &114,510  & 13.9 \\
  &                           &        & dev   &  47,016   & 28,115  & 12.8 \\
  &                           &        & test  &  46,559   & 28,350  & 12.7 \\
  \cmidrule(lr){3-7}
  &                           & \multirow{3}{*}{H3K79me3} & train & 326,286   &104,128  & 14.1 \\
  &                           &        & dev   &  37,459   & 23,783  & 13.0 \\
  &                           &        & test  &  35,128   & 21,888  & 12.2 \\
  \cmidrule(lr){3-7}
  &                           & \multirow{3}{*}{H3K9ac}   & train & 321,090   &104,878  & 14.5 \\
  &                           &        & dev   &  36,449   & 23,437  & 13.1 \\
  &                           &        & test  &  34,795   & 21,942  & 12.5 \\
  \cmidrule(lr){3-7}
  &                           & \multirow{3}{*}{H4}       & train & 174,731   & 74,297  & 15.0 \\
  &                           &        & dev   &  20,417   & 14,885  & 14.0 \\
  &                           &        & test  &  16,096   & 10,772  & 11.0 \\
  \cmidrule(lr){3-7}
  &                           & \multirow{3}{*}{H4ac}     & train & 384,137   &112,578  & 14.1 \\
  &                           &        & dev   &  43,842   & 26,847  & 12.9 \\
  &                           &        & test  &  43,287   & 26,715  & 12.7 \\
  \midrule

  \multirow[c]{3}{*}{Virus} & \multirow[c]{3}{*}{CVC}
    & \multirow{3}{*}{covid}   & train &  9,304,179 &  5,983  &126.9 \\
  &                         &        & dev   &  1,164,360 &  5,071  &127.0 \\
  &                         &        & test  &  1,162,628 &  5,214  &126.8 \\
\end{longtable}

\section{Full Results on GUE Benchmarks}
\label{app:full results}

\textcolor{black}{
We present the full comparison with baselines with various settings in Table \ref{tab:all compare}, where DNAZEN outperforms the baselines in most cases.
Following \cite{ji2021dnabert}, we use the following settings of the baseline models.
DNABERT (3‑mer) \citep{ji2021dnabert} is the original version of the DNABERT model, pretrained on human genomic data using overlapping 3‑nucleotide tokens to capture local sequence patterns. 
DNABERT (4‑mer), DNABERT (5‑mer), and DNABERT (6‑mer) follow exactly the same transformer architecture but swap in 4‑, 5‑, or 6‑nucleotide units, respectively. 
DNABERT‑2 \citep{ji2021dnabert} utilizes BPE tokenization and is pre-trained on the same dataset as DNAZEN. 
The Nucleotide Transformer (NT) \citep{dalla2024nucleotide} scales both data and model size: NT‑500M‑human is a 500 million‑parameter model trained solely on the GRCh38/hg38 dataset; 
NT‑500M‑1000g matches that size but learns from high‑coverage human genomes to expose population variation; 
NT‑2500M‑1000g grows to 2.5 billion parameters on the same 1000 Genomes corpus; 
and NT‑2500M‑multi preserves the 2.5 billion‑parameter scale while drawing training examples from 850 different species, enabling cross‑species generalization.
}

\begin{table}[t]
    \centering
    \caption{Comparison of DNAZEN and existing studies. \textcolor{black}{The DNABERT, DNABERT-2, and Nucleotide Transformer (NT) results are copied from \cite{zhou2023dnabert}}.}
    \label{tab:all compare}
    \setlength{\tabcolsep}{4pt}
    \begin{tabularx}{\textwidth}{@{\extracolsep{\fill}}l|cccccc}
        \toprule
        Task & \multicolumn{6}{c}{Epigenetic Marks Prediction} \\
        \midrule
        Dataset & H3    & H3K14ac & H3K36me3 & H3K4me1 & H3K4me2 & H3K4me3 \\
        \midrule
        DNABERT(3-mer)    & 74.15 & 42.07 & 48.49 & 42.95 & 31.34 & 28.92 \\
        DNABERT(4-mer)    & 73.03 & 41.88 & 48.03 & 41.06 & 30.66 & 25.31 \\
        DNABERT(5-mer)    & 73.40 & 40.68 & 48.29 & 40.65 & 30.67 & 27.10 \\
        DNABERT(6-mer)    & 73.10 & 40.06 & 47.25 & 41.44 & 32.27 & 27.81 \\
        DNABERT-2$^\dag$ & 80.17 & 57.42 & 61.90 & 53.00 & 39.89 & 41.20 \\
        NT-500M-human     & 69.67 & 33.55 & 44.14 & 37.15 & 30.87 & 24.06 \\
        NT-500M-1000g     & 72.52 & 39.37 & 45.58 & 40.45 & 31.05 & 26.16 \\
        NT-2500M-1000g    & 74.61 & 44.08 & 50.86 & 43.10 & 30.28 & 30.87 \\
        NT-2500M-multi    & 78.77 & 56.20 & 61.99 & 55.30 & 36.49 & 40.34 \\
        \midrule
        DNAZEN$^\dag$ & \underline{\textbf{81.09}} & \underline{\textbf{59.28}} & \underline{\textbf{63.19}} & \underline{\textbf{58.14}} & \underline{\textbf{39.73}} & \underline{\textbf{42.84}} \\
        \bottomrule
    \end{tabularx}
    \begin{tabularx}{\textwidth}{@{\extracolsep{\fill}}l|cccc|ccc}
        \toprule
         Task & \multicolumn{4}{c|}{Epigenetic Marks Prediction} & \multicolumn{3}{c}{Promoter Detection} \\
        \midrule
        Dataset & H3K79me3 & H3K9ac & H4 & H4ac & all & notata & tata \\
        \midrule
        DNABERT(3-mer)   & 60.12 & 50.48 & 78.27 & 38.60 & 90.44 & 93.61 & 69.83 \\
        DNABERT(4-mer)   & 59.77 & 51.44 & 78.28 & 36.40 & 89.54 & 92.65 & 66.78 \\
        DNABERT(5-mer)   & 59.61 & 51.11 & 77.27 & 37.48 & 90.16 & 92.45 & 69.51 \\
        DNABERT(6-mer)   & 61.17 & 51.22 & 79.26 & 37.43 & 90.48 & 93.05 & 61.56 \\
        DNABERT-2$^\dag$ & 65.46 & 57.07 & 81.86 & 50.35 & 88.31 & \textbf{94.34} & 68.79 \\
        NT-500M-human    & 58.35 & 45.81 & 76.17 & 33.74 & 87.71 & 90.75 & 78.07 \\
        NT-500M-1000g    & 59.33 & 49.29 & 76.29 & 36.79 & 89.76 & 91.75 & 78.23 \\
        NT-2500M-1000g   & 61.20 & 52.36 & 79.76 & 41.46 & 90.95 & 93.07 & 75.80 \\
        NT-2500M-multi   & 64.70 & 56.01 & 81.67 & 49.13 & \textbf{91.01} & 94.00 & \textbf{79.43} \\
        \midrule
        DNAZEN$^\dag$ & \underline{\textbf{67.78}} & \underline{\textbf{59.27}} & \underline{\textbf{82.24}} & \underline{\textbf{53.15}} & 87.03 & 93.80 & \underline{69.84} \\
        \bottomrule
    \end{tabularx}
    \begin{tabularx}{\textwidth}{@{\extracolsep{\fill}}l | ccccc | ccc}
        \toprule
         Tasks & \multicolumn{5}{c|}{Transcription Factor Prediction (Human)} & \multicolumn{3}{c}{Core Promoter Detection} \\
        \midrule
        Dataset & 0 & 1 & 2 & 3 & 4 & all & notata & tata \\
        \midrule
        DNABERT(3-mer)   & 67.95 & 70.90 & 60.51 & 53.03 & 69.76 & \textbf{70.92} & 69.82 & 78.15 \\
        DNABERT(4-mer)   & 67.90 & 73.05 & 59.52 & 50.37 & 71.23 & 69.00 & 70.04 & 74.25 \\
        DNABERT(5-mer)   & 66.97 & 69.98 & 59.03 & 52.95 & 69.26 & 69.48 & 69.81 & 76.79 \\
        DNABERT(6-mer)   & 66.84 & 70.14 & 61.03 & 51.89 & 70.97 & 68.90 & 70.47 & 76.06 \\
        DNABERT-2$^\dag$ & \textbf{69.12} & 71.87 & 62.96 & 55.35 & 74.94 & 67.50 & 69.53 & 76.18 \\
        NT-500M-human    & 61.59 & 66.75 & 53.58 & 42.95 & 60.81 & 63.45 & 64.82 & 71.34 \\
        NT-500M-1000g    & 63.64 & 70.17 & 52.73 & 45.24 & 62.82 & 66.70 & 67.17 & 73.52 \\
        NT-2500M-1000g   & 66.31 & 68.30 & 58.70 & 49.08 & 67.59 & 67.39 & 67.46 & 69.66 \\
        NT-2500M-multi   & 66.64 & 70.28 & 58.72 & 51.65 & 69.34 & 70.33 & \textbf{71.58} & 72.97 \\
        \midrule
        DNAZEN$^\dag$ & 68.35 & \underline{\textbf{75.07}} & \underline{\textbf{66.66}} & \underline{\textbf{60.68}} & \underline{{78.81}} & \underline{67.62} & 68.21 & \underline{\textbf{79.15}} \\
        \bottomrule
    \end{tabularx}
    \begin{tabularx}{\textwidth}{@{\extracolsep{\fill}}l | ccccc | c | c}
        \toprule
        Tasks & \multicolumn{5}{c|}{Transcription Factor Prediction (Mouse)} & \multicolumn{1}{c|}{Virus} & \multicolumn{1}{c}{Splice} \\
        \midrule
        Dataset & 0 & 1 & 2 & 3 & 4 & Covid & Reconstruct \\
        \midrule
        DNABERT(3-mer)   & 42.31 & 79.10 & 69.90 & 55.40 & 41.97 & 62.23 & 84.14 \\
        DNABERT(4-mer)   & 49.42 & 79.95 & 72.62 & 51.79 & 44.13 & 59.87 & 84.05 \\
        DNABERT(5-mer)   & 42.45 & 79.32 & 62.22 & 49.92 & 40.34 & 50.46 & 84.02 \\
        DNABERT(6-mer)   & 44.42 & 78.94 & 71.44 & 44.89 & 42.48 & 55.50 & 84.07 \\
        DNABERT-2$^\dag$ & \textbf{64.23} & \textbf{86.28} & 81.28 & 73.49 & 50.80 & 68.49 & 85.93 \\
        NT-500M-human    & 31.04 & 75.04 & 61.67 & 29.17 & 29.27 & 50.82 & 79.71 \\
        NT-500M-1000g    & 39.26 & 75.49 & 64.70 & 33.07 & 34.01 & 52.06 & 80.97 \\
        NT-2500M-1000g   & 48.31 & 80.02 & 70.14 & 42.25 & 43.40 & 66.73 & 85.78 \\
        NT-2500M-multi   & 63.31 & 83.76 & 71.52 & 69.44 & 47.07 & \textbf{73.04} & \textbf{89.35} \\
        \midrule
        DNAZEN$^\dag$ & 62.25 & 85.17 & \underline{\textbf{83.02}} & \underline{\textbf{75.12}} & \underline{\textbf{52.45}} & 66.54 & \underline{\textbf{89.35}} \\
        \bottomrule
    \end{tabularx}
\end{table}

\end{document}